\documentclass[10pt,twocolumn,letterpaper]{article}

\usepackage{iccv}
\usepackage{times}
\usepackage{epsfig}
\usepackage{graphicx}
\usepackage{amsmath}
\usepackage{amssymb}

\makeatletter
\def\thanks#1{\protected@xdef\@thanks{\@thanks
        \protect\footnotetext{#1}}}
\makeatother

\usepackage{multirow}
\usepackage{booktabs}
\usepackage{units}

\usepackage{amsmath}
\usepackage{amssymb}
\usepackage{mathtools}
\usepackage{amsthm}
\usepackage{enumitem}
\usepackage{bm}

\usepackage{xcolor}
\usepackage{chngcntr}

\usepackage[breaklinks=true,bookmarks=false,colorlinks,linkcolor=red]{hyperref}

\iccvfinalcopy 
\usepackage[accsupp]{axessibility}

\def\iccvPaperID{9458} 

\ificcvfinal\fi

\begin{document}

\title{Coarse-to-Fine Amodal Segmentation with Shape Prior}

\author{Jianxiong Gao$^{1}$, Xuelin Qian$^{1,\dagger}$, Yikai Wang$^{1}$, Tianjun Xiao$^{2,\dagger}$, Tong He$^{2}$, Zheng Zhang$^{2}$, Yanwei Fu$^{1}$
\thanks{$\dagger$: Co-corresponding authors.}\\
$^{1}$Fudan University, \quad $^{2}$Amazon Web Service\\
{\tt\small jxgao22@m.fudan.edu.cn, \{xlqian,yikaiwang19,yanweifu\}@fudan.edu.cn}\\ 
{\tt\small \{tianjux,htong,zhaz\}@amazon.com}}

\maketitle
\ificcvfinal\thispagestyle{empty}\fi

\begin{abstract}
Amodal object segmentation is a challenging task that involves segmenting both visible and occluded parts of an object. 
In this paper, we propose a novel approach, called Coarse-to-Fine Segmentation (C2F-Seg), that addresses this problem by progressively modeling the amodal segmentation.
C2F-Seg initially reduces the learning space from the pixel-level image space to the vector-quantized latent space.
This enables us to better handle long-range dependencies and learn a coarse-grained amodal segment from visual features and visible segments.
However, this latent space lacks detailed information about the object, which makes it difficult to provide a precise segmentation directly.
To address this issue, we propose a convolution refine module to inject fine-grained information and provide a more precise amodal object segmentation based on visual features and coarse-predicted segmentation.
To help the studies of amodal object segmentation, we create a synthetic amodal dataset, named as MOViD-Amodal (MOViD-A), which can be used for both image and video amodal object segmentation.
We extensively evaluate our model on two benchmark datasets: KINS and COCO-A. Our empirical results demonstrate the superiority of C2F-Seg. 
Moreover, we exhibit the potential of our approach for video amodal object segmentation tasks on FISHBOWL and our proposed MOViD-A. 
Project page at: \url{https://jianxgao.github.io/C2F-Seg}.

\end{abstract}

\begin{figure}[ht]
\begin{center}
\includegraphics[width=\columnwidth]{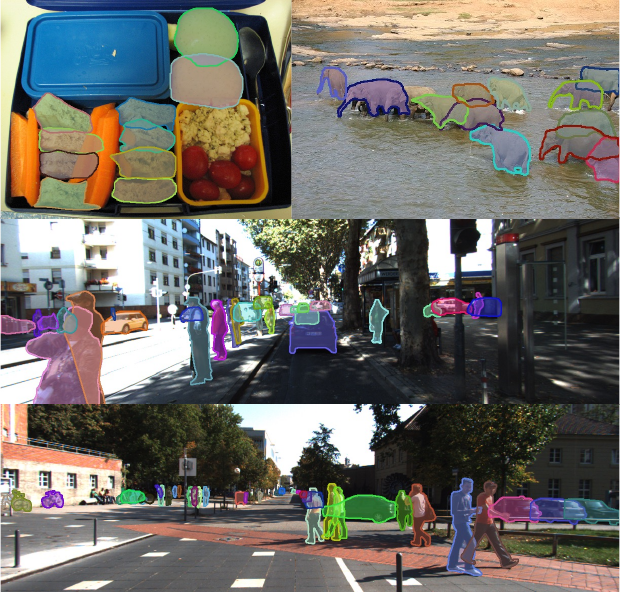}
\caption{
Visualization of predicted amodal masks in KINS and COCOA by C2F-Seg. Images in the top row are from COCOA, while the others are from KINS.
}
\label{fig:first_fig}
\end{center}
\vskip -0.3in
\end{figure}

\section{Introduction}

Amodal instance segmentation \cite{qi2019amodal} aims to extract complete shapes of objects in an image, including both visible and occluded parts. This task plays a vital role in various real-world applications such as autonomous driving \cite{qian2023impdet,geiger2012we}, robotics \cite{cheang2022learning}, and augmented reality \cite{park2008multiple,mathis2021fast}. For instance, in autonomous driving, partial understanding of the scene may result in unsafe driving decisions.

Typically, existing approaches \cite{mohan2022amodal,tran2022aisformer,xiao2021amodal,ke2021deep,follmann2019learning} build new modules on the detection framework, by additionally introducing an amodal branch that predicts the complete mask perception of the target object. The central idea lies in imbibing a holistic understanding of shape (\ie, shape prior) through multi-task learning by harnessing the supervised signals of the visible and full regions. While these approaches have yielded promising outcomes in recent years, the task of amodal segmentation remains fraught with challenges. One of the main challenges of amodal segmentation is that it is an ill-posed problem, meaning that there are many non-unique and reasonable possibilities for perceiving occluded areas, particularly for elastic bodies like people and animals. On the other hand, there are intricate categories and shapes of objects in real-world scenarios, which would pose significant challenges to prior learning of shapes.

In this paper, we advocate that shape priors are essential for amodal segmentation, since the shape of an object is usually determined by its function, physiology, and characteristics. For example, a carrot has a long shape, while an apple has a round shape. Thus, the potential distribution of this object can be learned via neural networks. 
Nevertheless, we argue that while shape prior can only provide a basic outline and may not capture individual differences or highly local information. Meanwhile, it is possible for the shape prior to being inconsistent with the observed visible area due to factors like pose and viewpoint.
To this end, we in this paper propose to generate amodal segments progressively via a coarse-to-fine manner.
Specifically, we divide the segmentation of amodal object into two phases: a coarse segmentation phase where we use the shape prior to generate a plausible amodal mask, and a refinement phase
is adopted to refine the coarse amodal mask to get the precise segmentation.

In the coarse segmentation phase, as we only need to provide a coarse mask, we perform the segmentation in a low-dimension vector-quantized latent space to reduce the learning difficulty and accelerate the inference process.
The segmentation in such latent space is resorted to the popular mask prediction task adopted in BERT~\cite{kenton2019bert} and MaskGIT~\cite{chang2022maskgit}.
Specifically, we adopt a transformer model which takes as inputs the ResNet visual feature, the vector-quantized visible segments, and the ground-truth amodal segments masked in a high ratio.
Then the transformer is trained to reconstruct the masked tokens of the amodal segments.
This mask-and-predict procedure~\cite{chang2022maskgit} leads to natural sequential decoding in the inference time.
Starting with an all-mask token sequence of amodal segments, our transformer gradually completes the amodal segments. Each step increasingly preserves the most confident prediction.

In the second refinement phase, our model learns to inject details to the coarse-prediction and provide a more precise amodal object segmentation.
Our convolutional refinement module takes as inputs the coarse-predicted segments and the visual features.
Imitating the human activity for visual stimulus, we construct a semantic-inspired attention module as an initial stimulus, and then gradually inject the visual features to the segments through convolution layers.

With this coarse-to-fine architecture design, our C2F-Seg complements the latent space that is easier-to-learn, the transformer that has superiority of long-range dependency, and the convolutional model that can supplement details, and results in a better amodal object segmentation.
Our framework is flexible to generalize to video-based amodal object segmentation tasks.
Guided by the shape prior and visual features of related frames, our model can generate precise amodal segments, and is even capable of generating amodal segments when the object is totally invisible, as shown in Figure~\ref{fig:maskvit_visualization}.

In order to evaluate the performance of our C2F-Seg. We conduct experiments both on image and video amodal segmentation benchmarks. For image amodal segmentation, our model reaches 36.5/36.6 on AP, 82.22/80.27 on full mIoU and 53.60/27.71 on occluded mIoU for KINS and COCOA respectively. For video amodal segmentation, our model reaches 91.68/71.30 on full mIoU and 81.21/36.04 on occluded mIoU for FISHBOWL and MOViD-A respectively. C2F-Seg outperforms all the baselines and achieves state-of-the-art performance.

Our contributions can be summarized as:
\begin{itemize}[leftmargin=*,itemsep=0pt,topsep=0pt,parsep=0pt]
\item We propose a novel coarse-to-fine framework, which consists of a mask-and-predict transformer module for coarse masks and a convolutional refinement module for refined masks. It imitates human activity and progressively generates amodal segmentation, mitigating the effect of detrimental and ill-posed shape priors.
\item We build a synthetic dataset MOViD-A for amodal segmentation, which contains 838 videos and 12,299 objects. We hope it will advance research in this field. We release the dataset on our project page.
\item Extensive experiments are conducted on two image-based benchmarks, showing the superiority of our methods over other competitors. Moreover, our framework can be easily extended to video-based amodal segmentation, achieving state-of-the-art performance on two benchmarks.
\end{itemize}

\section{Related Works}
\label{Related Works}

\begin{figure*}
    \centering
    \includegraphics[width=17cm]{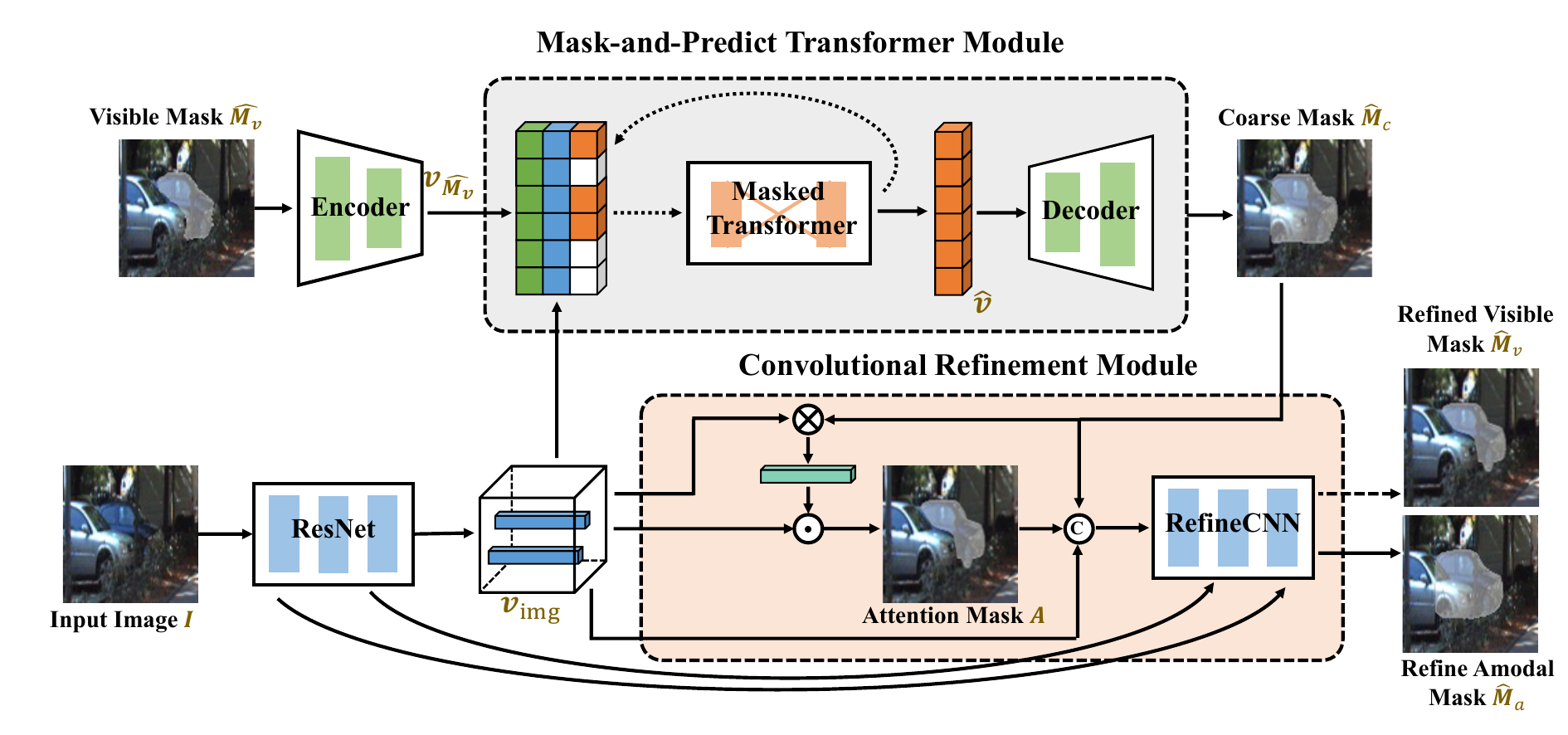}
\caption{Illustration of our C2F-Seg framework. 
C2F-Seg first generates a coarse mask from the visible mask and visual features via the mask-and-predict procedure with transformers.
Then this coarse amodal mask is refined with a convolutional module guided by human-imitated attention on visual features of the amodal object.
The learning of visible mask is used as an auxiliary task in training, while in inference we only provide an estimation of amodal mask.
}
    \label{fig:architecture}
\end{figure*}

\noindent \textbf{Amodal Instance Segmentation}~\cite{zhu2017semantic} is a challenging task that involves predicting the shape of occluded objects in addition to the visible parts.
To enhance the learning of such connection between the amodal segments and the category label, or prior, previous approaches design specific architectures.
MLC~\cite{qi2019amodal} learns the visible masks and amodal masks separately via two network branches.
AISFormer~\cite{tran2022aisformer} enhances the extraction of the long-range dependency via transformers, and utilizes multi-task training to learn a more comprehensive segmentation model.
VRSP~\cite{xiao2021amodal} for the first time explicitly designs the shape prior module to refine the amodal mask.
There are also some approaches~\cite{zhu2017semantic, follmann2019learning, zhang2019learning, xiao2020amodal, zhan2020self, tangemann2021unsupervised,ke2021occlusion,yang2019embodied, sun2022amodal, ling2020variational} focus on modeling shape priors with shape statistics, making it challenging to extend their models to open-world applications where object category distributions are long-tail and hard to pre-define.
SaVos~\cite{yao2022self} leverages spatiotemporal consistency and dense object motion to alleviate this problem. 
However, SaVos requires additional knowledge of optical flow known to cause object deformation in the presence of camera motion.
In contrast, our method doesn't need the optical flow anymore. We propose a new framework to learn generic object prior in vector-quantized latent space with transformer to predict the coarse amodal masks of occluded objects. Then we use a CNN-based refine module to polish up the coarse mask in pixel-level to get the fine amodal mask.

\noindent \textbf{Vision Transformer}.
The self-attention module~\cite{vaswani2017attention} has enabled impressive performance in various natural language processing and vision tasks through transformer-based methods such as BERT~\cite{devlin2018bert} and ViT~\cite{dosovitskiy2020image}, specifically in vision tasks such as image classification~\cite{dosovitskiy2020image}, object detection~\cite{carion2020end}, image/video synthesis~\cite{chang2022maskgit,gupta2022maskvit}.
Nevertheless, applying transformers to autoregressively generate high-resolution images is computationally expensive and memory-intensive~\cite{chen2020generative,kumar2021colorization}. 
Thus, new techniques like dVAE~\cite{ramesh2021zero} and VQ-VAE~\cite{oord2017neural} have been developed to represent images as discrete codes and shorten the sequence. VQ-GAN builds on VQ-VAE~\cite{esser2021taming}  by using GANs to improve efficiency, but both methods still use a single codebook to quantize the entire image.
Amodal segmentation can also benefit from transformer adaptation.
AISformer~\cite{tran2022aisformer} employs transformer-based mask heads to predict amodal masks, following the approach of DETR~\cite{carion2020end}. 
Our framework adopts transformer to utilize the mask-and-predict formulation of amodal object segmentation, inspired by MaskGIT and MaskViT~\cite{chang2022maskgit, gupta2022maskvit}. Then, we refine the prediction using a CNN-based module for precise segmentation.

\section{Coarse-to-Fine Segmentation}
\label{Methods}

\subsection{Problem Setup}
Amodal object segmentation aims to segment not only the visible parts of an object but also its occluded parts.
Formally, amodal object segmentation takes as inputs an input image $\bm{I}$, a bounding box of the Region-of-Interest (ROI).
The amodal object is only partially visible in the image, decomposed into visible parts and occluded parts.
The visible parts can be segmented via standard segmentation algorithms, but the invisible occluded parts needs to be estimated rather than segmented.
Following~\cite{yao2022self}, 
we denote the visible segment as $\bm{M}_v$ and the full/amodal segment as $\bm{M}_a$ such that $\bm{M}_a$ consisted of both the visible segment and the invisible segment.
Thus our target is to estimate $\bm{M}_v$ and $\bm{M}_a$ simultaneously from the ROI of $\bm{I}$.

We  utilize current segmentation algorithms to provide an estimation of visible segment $\hat{\bm{M}}_v$.
Then based on $\bm{I}$ and $\hat{\bm{M}}_v$, we construct our C2F-Seg framework by two stages.
In the first stage, we estimate a coarse-grained segment $\bm{M}_c$ based on the vector-quantized latent space by transformer.
Then we adopt a convolutional module to refine the estimation and provide a precise fine-grained prediction $\hat{\bm{M}}_a$ as the final estimation of $\bm{M}_a$.
In the following, we introduce each component of our C2F-Seg framework in details.

\subsection{Vector-Quantized Latent Space}

Our latent space is inspired by the well-known VQ-GAN~\cite{esser2021taming}.
Specifically, we adopt an encode-decode architecture with encoder $E$ and decoder $D$ with convolutional layers.
For input mask $\bm{M}\in\mathbb{R}^{H\times W}$, the encoder projects it to the continuous latent code $\hat{\bm{z}} = E(\bm{M})$ from a learned, discrete codebook $\mathcal{Z}=\{z_k\}^K_{k=1}\subset \mathbb{R} ^{n_z}$, where $n_z$ is the dimension of codes.
Then the closest codebook entry of each spatial code $z_{ij}$ is utilized to get a discrete representation from the codebook vocabulary
\begin{equation}
z_{\mathbf{q}}=\mathbf{q}(\hat{z}) :=\left(\underset{z_k \in \mathcal{Z}}{\arg \min }\left\|\hat{z}_{i j}-z_k\right\|\right) \in \mathbb{R}^{h \times w \times n_z},
\end{equation}
where $z_k$ means the closest codebook entry of each spatial code $z_{ij}$. 
With this discrete representation, the decoder $D$ reconstructs the input mask $\bm{M}$ as 
\begin{equation}
\hat{\bm{M}}\coloneqq D\left(z_{\mathbf{q}}\right)=D(\mathbf{q}(E(\bm{M}))).
\end{equation}
With properly trained encoder and decoder, we can get a latent representation $s \in\{0, \ldots,|\mathcal{Z}|-1\}^{h \times w}$ in terms of the codebook-indices for each mask, which consists the latent space on which our first learning stage performs. We initialize an embedding for the indices $s$ as the input to the transformer model:
\begin{equation}
\bm{v}_{\bm{M}}\coloneqq \mathrm{Embed}(s).
\end{equation}

\begin{figure}[t]
\begin{center}
\centerline{\includegraphics[width=\columnwidth]{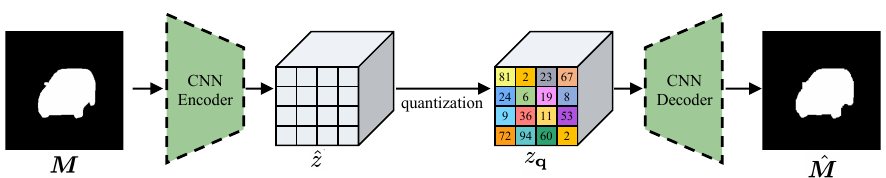}}
\caption{The architecture of Vector-Quantization model. The trained latent representation of masks are used in our transformer.}
\end{center}
\vskip -0.3in
\end{figure}

While it is common to utilize a VQ-GAN to encode the input image in the corresponding latent space, our preliminary experiments reveal a decrease in performance when employing this method. This could be due to the fact that current approaches utilizing the VQ-GAN for image encoding and latent space learning are typically reliant on a vast training dataset comprising millions, if not billions, of data points.
Unfortunately, for amodal object segmentation tasks, we have access to only a limited training set, which may not be sufficient to train a potent embedding. As a result, in practical scenarios, we resort to utilizing a pretrained ResNet to extract and flatten the visual features of the input image as the transformer model input:

\begin{equation}
\bm{v}_{\mathrm{img}}\coloneqq\mathrm{Flatten}(\mathrm{ResNet}(\bm{I})).
\end{equation}
By adopting this approach, we can alleviate the learning complexity and enhance the segmentation ability. Since the embedding of masks is initialized randomly, we choose to set the embedding dimension to match the size of the visual features for improved alignment.

\subsection{Mask-and-Predict Transformer}
Having established the latent space's architecture, as described in the preceding subsection, we now possess an image representation denoted as $\bm{v}_{\mathrm{img}}$ and a visible segment representation referred to as $\bm{v}_{\hat{\bm{M}}v}$. Our aim is to predict the amodal object segmentation, denoted as $\bm{v}_{\bm{M}_a}$.

To achieve this, we introduce a [MASK] token apart from the learned mask codebook.
Then we initialize $\hat{\bm{v}}$ as all [MASK] tokens with the same dimension of $\bm{v}_{\hat{\bm{M}}_v}$.
Then we concatenate $\bm{v}_{\mathrm{img}}, \bm{v}_{\hat{\bm{M}}_v}$, and $\hat{\bm{v}}$ as the input of the transformer model.

The training objective of the transformer model is to minimize the negative log-likelihood as
\begin{equation}
\mathcal{L}\coloneqq -\mathbb{E}\left[\sum_{i}\log p(\bm{v}_{\bm{M}_a,i}\mid \hat{\bm{v}}, \bm{v}_{\mathrm{img}}, \bm{v}_{\hat{\bm{M}}_v})\right].
\end{equation}

Nevertheless, learning to make one-step prediction are known to be challenging. Therefore, we draw inspiration from the general concept behind mask-and-predict approaches such as BERT~\cite{kenton2019bert},  MaskGIT~\cite{chang2022maskgit} and MaskVIT~\cite{gupta2022maskvit}. By doing so, we can simplify the objective by masking specific codes within the amodal segment representation, and then predicting the masked portions. 
Denote the masking operator as $\mathcal{M}$, our training objective now becomes
\begin{equation}
\mathcal{L}\coloneqq -\mathbb{E}\left[\sum_{i}\log p(\bm{v}_{\bm{M}_a,i}\mid \mathcal{M}(\bm{v}_{\bm{M}_a}), \bm{v}_{\mathrm{img}}, \bm{v}_{\hat{\bm{M}}_v})\right].
\end{equation}

Developing an appropriate masking policy is critical to the overall approach's success. If we only mask a negligible fraction of the amodal segments, the task becomes simplistic and fails to generalize to testing stages. Conversely, if we mask a substantial portion, it may prove too challenging to learn. To address this, we uniformly select the masking ratio from 50\% to 100\% in practical scenarios. This approach enables us to strike a balance between learning difficulty and training-testing consistency.

During inference, we take iterative inference method to complete the amodal masks in $K$ steps. At each step, our model predicts all tokens simultaneously but only keeps the most confident ones. The remaining tokens are masked out and re-predicted in the next iteration. The mask ratio is made decreasing until all tokens are generated within T iterations.

After estimating $\hat{\bm{v}}$, we use the decoder $D$ to reconstruct the coarse-predicted amodal mask 
\begin{equation}
\hat{\bm{M}}_c = D(\hat{\bm{v}}).
\end{equation}

\subsection{Convolutional Refinement}
Although we train the VQ-GAN model to reconstruct the mask as precisely as possible, it inevitably losses some details of the mask and thus is only a coarse estimation.
To recover these details, we adopt a convolutional refinement module.

Our convolutional refinement module takes as inputs the image features $\bm{v}_{\mathrm{img}}$ and the estimated coarse amodal mask $\hat{\bm{M}}_c$.
We first downsample the coarse amodal mask to match the dimension of the ResNet features, yielding downsampled $\hat{\bm{M}}_{cd}$.
Note that in our mask-and-predict transformer, we adopt a vector-quantization module to align mask and visual features.
However, this alignment requires extra training.
Thus in our convolutional refinement module, we direct downsample the mask to avoid additional training for efficiency.
As the mask can be regarded as a hard attention map, we directly encourage an attention on the amodal object via
\begin{equation}
\bm{A}\coloneqq\mathrm{softmax}(\frac{\hat{\bm{M}}_{cd}\bm{v}_{\mathrm{img}}^{\top}}{\sqrt{d}})\odot\bm{v}_{\mathrm{img}},
\end{equation}
where $\odot$ is the element-wise multiplication.

Then the convolutional refinement module learns to predict the visible segment and amodal segment simultaneously
\begin{equation}
\hat{\bm{M}}_a, \hat{\bm{M}}_v = \mathrm{Conv}(\bm{v}_{\mathrm{img}},\bm{A},\hat{\bm{M}}_{cd}).
\end{equation}
The convolutional refinement is trained to minimize binary cross-entropy loss for visible mask and amodal mask simultaneously.
\begin{equation}
\mathcal{L}_r\coloneqq \mathrm{BCE}(\hat{\bm{M}}_a, \bm{M}_a) + \mathrm{BCE}(\hat{\bm{M}}_v, \bm{M}_v).
\end{equation}

\subsection{Extension to Video Amodal Segmentation} 
\begin{figure}[t]
\vskip -0.1in
\begin{center}
\centerline{\includegraphics[width=\columnwidth]{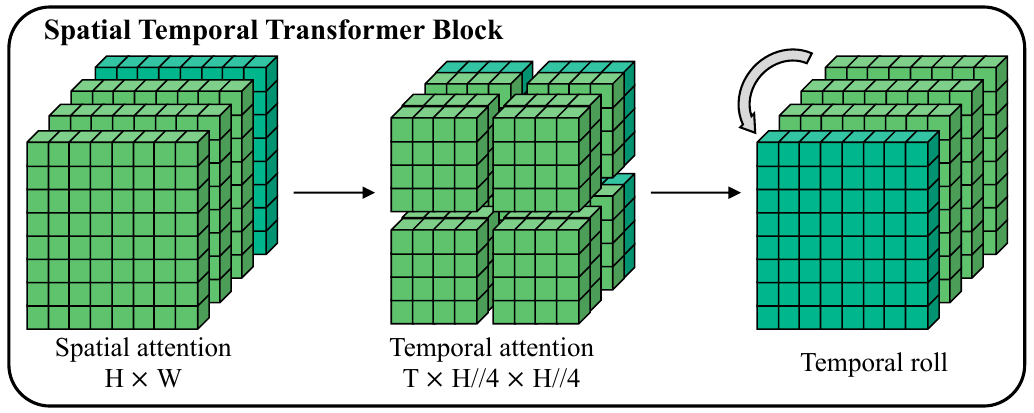}}
\caption{Architecture of Spatial Temporal Transformer Block(STTB). For video tasks, we roll the features in temporal dimension after each transformer block and recover the normal order at the end of our model.}
\label{stlayers}
\end{center}
\vskip -0.4in
\end{figure}

Our framework can generalize to video amodal object segmentation easily.
Specifically, our model leverages temporal-spatial attention \cite{gupta2022maskvit} to capture the temporal features throughout the entire video and model the amodal masks. Each transformer block comprises a spatial layer and a temporal layer, as shown in Figure~\ref{stlayers}. The spatial layer functions similar to classical self-attention layers, while the temporal layer splits the codes in the spatial dimension and stacks them into a $(T,\; h,\; w)$ size in the temporal dimension. It then performs self-attention on these codes to capture the temporal relationships in the video.

Compared with object in a single image, object in video suffers from occlusion of different parts in different frames, and the object itself may also undergo rotation and deformation relative to the camera. Therefore, it is essential to enhance the spatial-temporal modeling ability of our model to accurately learn the complete shape of the target objects. To fully extract spatial-temporal information, our model rolls the features in the temporal dimension by $\nicefrac{T}{2}$ frames after each transformer block. This operation significantly improves the performance, as discussed in the supplementary.

\begin{figure}[t]
\begin{center}
\includegraphics[width=\columnwidth]{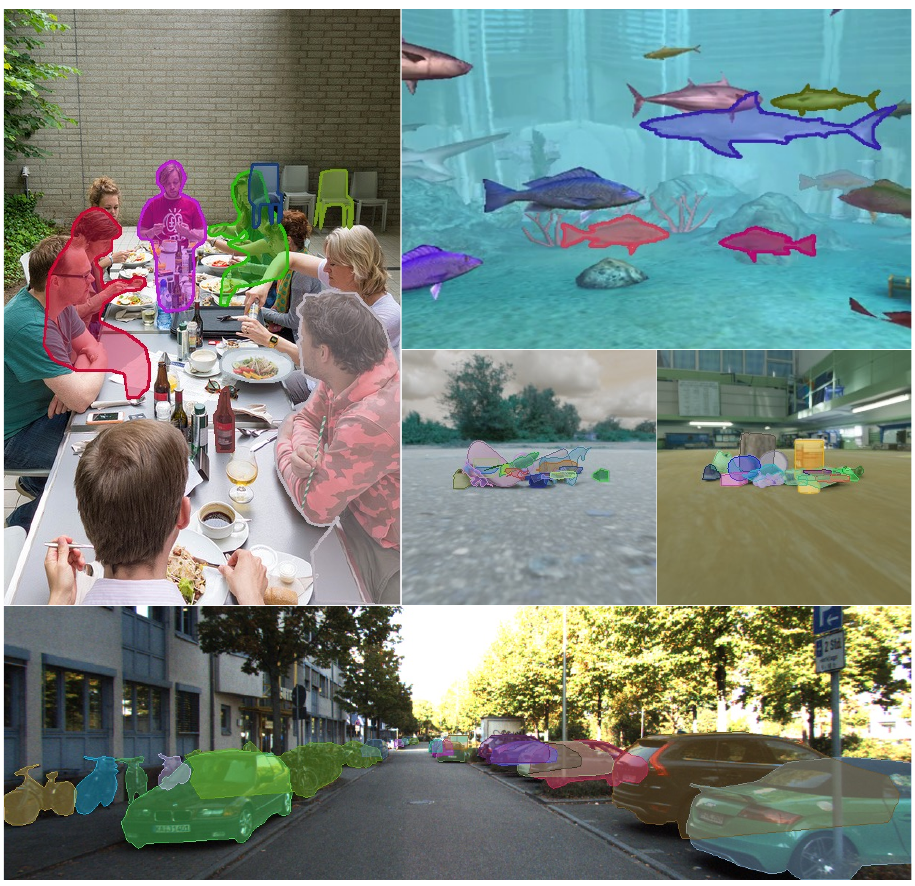}
\caption{Overview of all the datasets used in our paper. The image at the top left panel is selected from COCOA. The top right panel shows an example of Fishbowl. Below are two images from MOViD-A. The bottom panel is an image from KINS.}
\label{fig:dataset_example}
\end{center}
\vskip -0.3in
\end{figure}

\section{Experiments} \label{experiments}

\begin{figure*}[ht]
\begin{center}
\includegraphics[width=16cm]{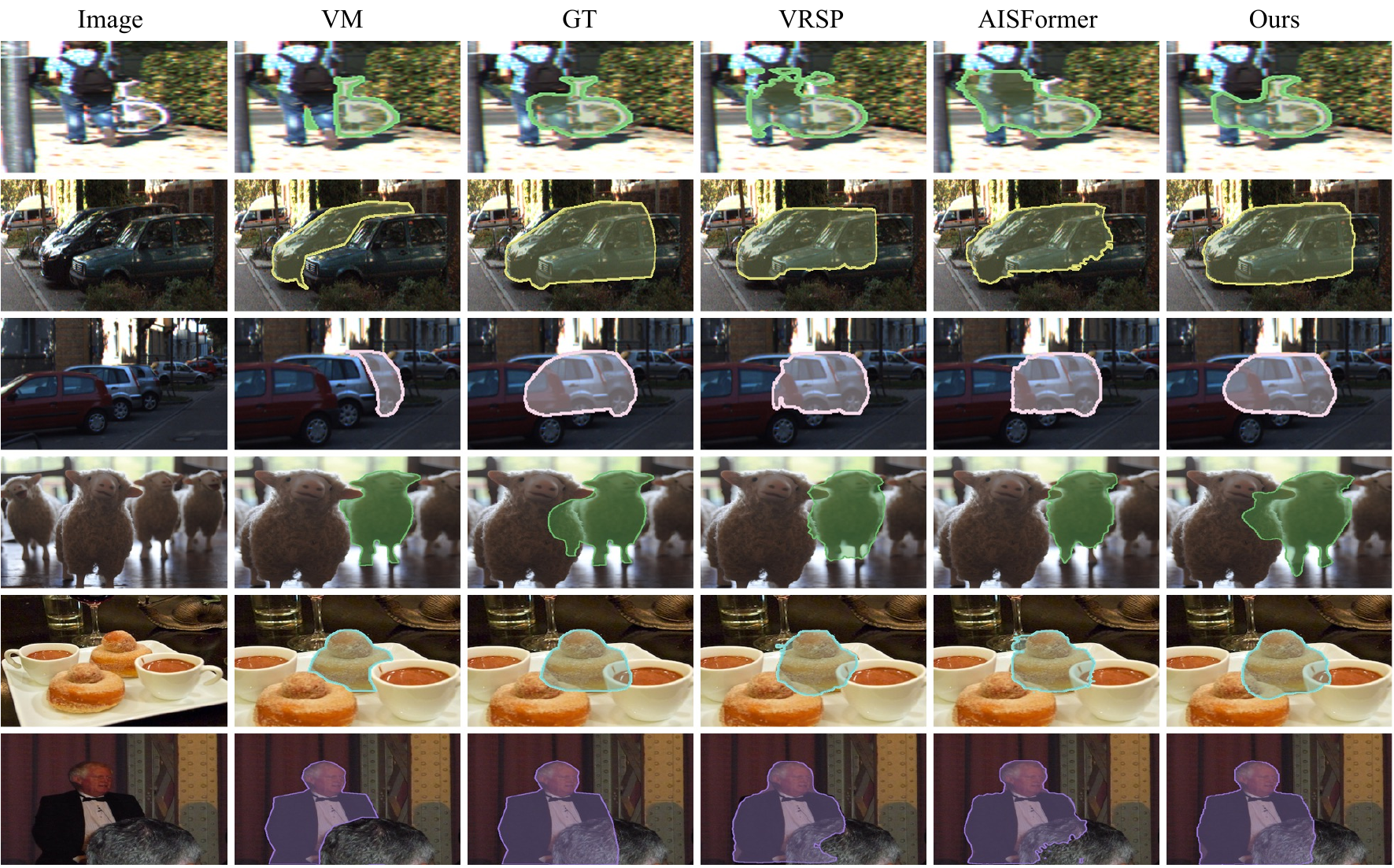}
\caption{The qualitative results estimated by VRSP, AISFormer, and our method.
VM and GT indicate ground-truth visible mask and amodal mask, respectively.
}
\label{fig:vi_img_results} 
\end{center}
\vskip -0.12in
\end{figure*}

\begin{table*}[ht]  \small 
\centering{
    \setlength{\tabcolsep}{2mm}{
    \begin{tabular}{l|cccccc|cccccc}
    \toprule
    \multicolumn{1}{c|}{\multirow{2}{*}{\textsc{Methods}}} & \multicolumn{6}{c|}{KINS} & \multicolumn{6}{c}{COCOA}\\
        & $AP$ & $AP_{50}$ & $AP_{75}$ & $AR$ &  mIoU$_{full}$ &  mIoU$_{occ}$ & $AP$ & $AP_{50}$ & $AP_{75}$ & $AR$ &  mIoU$_{full}$ &  mIoU$_{occ}$ \\
    \midrule
    \midrule
    PCNet~\cite{zhan2020self}  & 29.1 & 51.8         &  29.6     & 18.3 & 78.02 & 38.14
    & - & - & - & - & 76.91 & 20.34
    \\
    Mask R-CNN~\cite{he2017mask}  & 30.0 & 54.5 & 30.1      & 19.4 & - & - 
    & 28.0 & 53.7 & 25.4 & 29.8 & - & - \\
    ORCNN~\cite{follmann2019learning}  & 30.6 & 54.2      & 31.3      & 19.7 & - & - 
    & 28.0 & 53.7 & 25.4 & 29.8 & - & - \\
    VRSP~\cite{xiao2021amodal}        & 32.1 & 55.4  & 33.3   & 20.9 & 80.70 & 47.33
    & 35.4 & 56.0 & \textbf{38.7} & 37.1 & 78.98          & 22.92 \\
    AISformer~\cite{tran2022aisformer}$^{\dag}$   & 33.8 & 57.8  & 35.3   & 21.1 & 81.53 & 48.54 
    & 29.0 & 45.7 & 31.8 & 31.1 & 72.69          & 13.75\\
    \hline
    C2F-Seg (\textit{ours})    & \textbf{36.5}&  \textbf{58.2}  & \textbf{37.0} &  \textbf{22.1}&  \textbf{82.22} & \textbf{53.60}
     & \textbf{36.6}  & \textbf{57.0} & 38.5  & \textbf{38.5 }& \textbf{80.28} &\textbf{27.71} \\
    \bottomrule
    \end{tabular}}}
\caption{
\textbf{Performance comparison on the KINS and COCOA.}
We fine-tune the AISformer (marked by $\dag$) on COCOA from the official model trained on KINS.
Other results are reported in AISformer.
\label{tab:metric_for_KINS_and_cocoa}
}
\vskip -0.05in
\end{table*}

\noindent \textbf{Datasets.} 
To evaluate the efficacy of our proposed model, we conduct comprehensive experiments on both image and video amodal segmentation benchmarks. 
\textbf{1) KINS} \cite {qi2019amodal} is a large-scale amodal instance dataset, which is built upon KITTI \cite{geiger2012we}. It contains 7 categories that are common on the road, including car, truck, pedestrian, \textit{etc}. There are 14,991 manually annotated images in total, 7,474 of which are used for training and the remaining for testing.
\textbf{2) COCOA} \cite{zhu2017semantic} is derived from COCO dataset \cite{lin2014microsoft}. It consists of 2,476 images in the training set and 1,223 images in the testing set. There are 80 objects in this dataset.
\textbf{3) FISHBOWL} \cite{tangemann2021unsupervised} is a video benchmark, recorded from a publicly available WebGL demo of an aquarium\cite{WebGLfish}. Following \cite{yao2022self}, we select 10,000 videos for training and 1,000 for testing, each with 128 frames.
\noindent \textbf{4) MOViD-A} is a video-based synthesized dataset. We create it from MOVi dataset \footnote{\url{https://github.com/google-research/kubric/tree/main/challenges/movi}} for amodal segmentation. The virtual camera is set to go around the scene, capturing about 24 consecutive frames. We randomly place $10 \sim 20$ static objects that heavily occlude each other in the scene. Finally, we collect 630 and 208 videos for training and testing. Examples are shown in Figure~\ref{fig:dataset_example}.

\noindent \textbf{Metrics.} 
For evaluation, we adopt standard metrics as in most amodal segmentation literature~\cite{ke2021deep,xiao2021amodal,tran2022aisformer}, namely mean average precision (AP) and mean average recall (AR). Furthermore, We use mean-IoU~\cite{qi2019amodal,yao2022self} to measure the quality of predicted amodal masks. It is calculated against the ground-truth amodal mask (mIoU$_{full}$) or the occluded region (mIoU$_{occ}$). Occluded mIoU measures the complete quality of the occluded part of target objects directly. It is worth noting that occluded mIoU is a crucial indicator for admodal segmentation. Following \cite{yao2022self}, we specially only compute mIoU for objects on FISHBOWL with the occlusion rate from 10 to 70\%, and all detected objects on other datasets are involved for evaluation.

\begin{figure*}[t]
\begin{center}
\centerline{
\includegraphics[width=17cm]{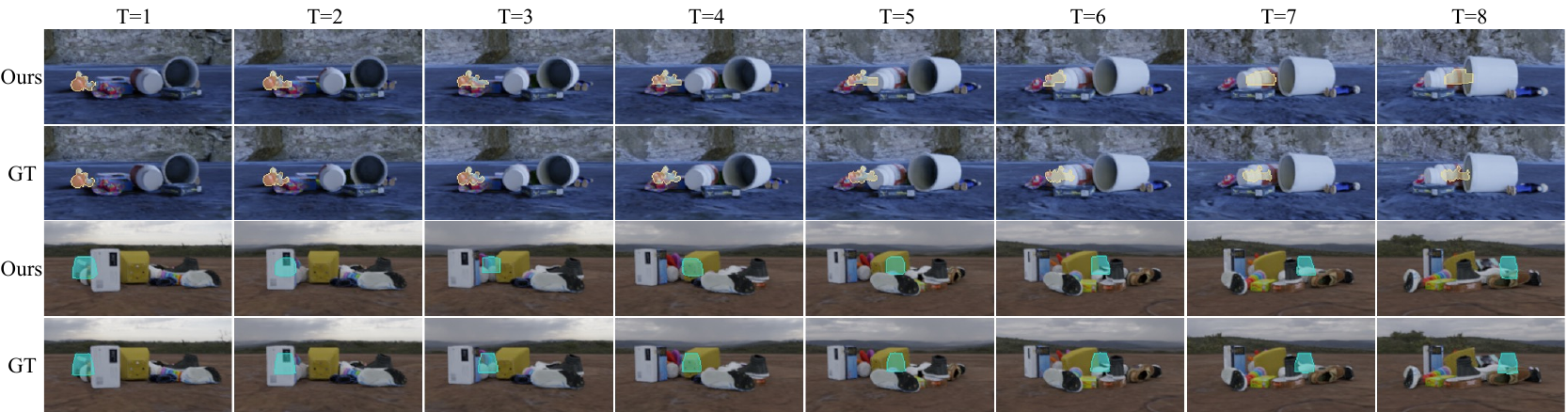}
}
\caption{The visulazation results of C2F-Seg on video dataset. The orange toy and the cyan box are invisible in a few frames, our model still produces approximate complete amodal masks. Best viewed in color and zoom in.}
\label{fig:maskvit_visualization}
\end{center}
\vskip -0.3in
\end{figure*}

\noindent\textbf{Implementation Details}.
Our framework is implemented on PyTorch platform. 
Considering that competitors for image-based amodal segmentation all include a detection branch, we use pre-detected visible bounding boxes and masks by AISFormer \cite{tran2022aisformer}, for fair comparison. Particularly,  since \cite{tran2022aisformer} does not provide weights on COCOA, we turn back to use the model trained by VRSP \cite{xiao2021amodal}. For video benchmarks, all baselines and our model take ground-truth visible masks as inputs.
We use bounding boxes of visible regions, which enlarge 2 times, to crop images and  masks as inputs. The inputs are all resized to $256 \times 256$. For data augmentation, morphology dilation, erosion and Gaussian blur are applied to mask inputs. AdamW optimizer~\cite{loshchilov2017decoupled} with a learning rate of 3e-4 is adopted for all experiments. We train the model using a batch size of 16 for all datasets except MOViD-A, which has a batch size of 24. The total number of iterations set for KINS, COCOA, FISHBOWL and MOViD-A datasets is 45K, 10K, 75K and 75K, respectively. For architecture design, we set the number of transformer layers is 12, and the feature dimension is 768 for image dataset. When training for FISHBOWL and MOViD-A, we adjust the number of transformer layers to 8, for we adjust the transformer block into spatial temporal one. For the codebook size of vector-quantized latent space, we set 256 for all the datasets.
The vector-quantization model is trained for each dataset, respectively. During inference, we consistently set the iterative step $K$ to 3 to demonstrate the generalizability of our method.

\subsection{Results of Image Amodal Segmentation}
We first compare our C2F-Seg with several image-based competitors on KINS and COCOA dataset. As shown in Table~\ref{tab:metric_for_KINS_and_cocoa}, we report results of both AP and mIoU. From the table, we can observe that
(1) our model achieves state-of-the-art performance on both datasets across most of metrics. For KINS, we outperform the second-best method by a margin of at least 5 points on mIoU$_{occ}$. 
Despite COCOA being a more challenging dataset than KINS due to its diverse object categories and intricate shapes, our method still yields better results compared to other approaches.
It clearly suggests the superiority of our proposed method.
(2) Compared with AISFormer \cite{tran2022aisformer} that also utilizes transformers for amodal segmentation, we beat it by 2.7\% and 1.0\% on AP and AR metrics. Moreover, VRSP~\cite{xiao2021amodal} utilizes learned shape prior to refine the predict amodal mask. Differently, our C2F-Seg leverages shape prior to obtain coarse amodal region and further complete it with visual features. As expected, we achieve 1.4\% and 4.79\% higher results on AR and mIoU$_{occ}$, which significantly shows the advantage of our design.

Qualitative results estimated by VRSP, AISFormer, and our method are further illustrated in Figure~\ref{fig:vi_img_results}. As observed, our method can segment more occluded regions with accurate shapes, owing to the help of excellent shape prior and precise refine module.
For visualizations from the 2nd to 4th row, the predictions of our method are not misled by occlusions which have the same category as target objects, especially when the occlusion rate is very large or relatively small. 
For objects that have intricate and delicate contours, such as the bicycle in 1st row, both VRSP and AISFormer fail to precisely segment the occluded area of rear-wheel and the visible region of saddle. By comparison, our method has shown to be successful in tackling this challenging case, and delivers good performance.

\begin{table} \small
\centering
\setlength{\tabcolsep}{1.5mm}{
{\begin{tabular}{lcccc}
\toprule
\multicolumn{1}{c}{\multirow{2}{*}{\textsc{Methods}}} & \multicolumn{2}{c}{FISHBOWL} & \multicolumn{2}{c}{MOViD-A} \\
 & mIoU$_{full}$ & mIoU$_{occ}$ & mIoU$_{full}$ & mIoU$_{occ}$  \\
\midrule
\midrule
\textit{visible masks}     & 68.53   &   -   & 56.92  & - \\
\hline 
Convex & 77.61 & 46.38 & 60.18  & 16.48 \\
PCNET~\cite{zhan2020self}  & 87.04 & 65.02 & 64.35  & 27.31 \\
SaVos~\cite{yao2022self}  & 88.63 & 71.55 & 60.61  & 22.64 \\
AISformer~\cite{tran2022aisformer}  & - & - & 67.72  & 33.65 \\
\hline
C2F-Seg (\textit{ours})   & \textbf{91.68} & \textbf{81.21} & \textbf{71.67} & \textbf{36.13}  \\
\bottomrule
\end{tabular}}}
\caption{
\textbf{Quantitative results on FISHBOWL and MOViD-A.} We report and compare the Mean-IoU metrics for FISHBOWL and MOViD-A of C2F-Seg with baselines.
\label{tab:fishbowl_MOVi_amodal}
}
\vskip -0.15in
\end{table}

\subsection{Results of Video Amodal Segmentation}
\label{subsec:video_results}
We further investigate the efficacy of our model for video amodal segmentation task. 
Table \ref{tab:fishbowl_MOVi_amodal} shows the mean-IoU metrics of C2F-Seg and baselines on FISHBOWL and MOViD-A datasets. Importantly, our method outperforms all the baselines, getting 81.21 and 36.04 results on occluded mIoU. Particularly, 
we achieve 4.64/16.19 and 3.05/9.66 higher performance than PCNET~\cite{zhan2020self} and SaVos~\cite{yao2022self} on FISHBOWL, respectively. On MOViD-A, we also achieve 10.69/13.4, 3.58/2.39 higher performance than SaVos and AISformer~\cite{tran2022aisformer}, respectively.
It is worth noting that MOViD-A presents much more challenges, including multiple objects, lens distortion, and change of view point. 
Nevertheless, our model remains effective in handling these challenging scenarios. 
In addition, we provide the lower bound by directly evaluating with input visible masks. 
All of these observations strongly demonstrate the effectiveness of our proposed model, highlighting its generalization ability to video amodal segmentation task. 
Figure~\ref{fig:maskvit_visualization} shows the qualitative results of our model for MOViD-A. For the extreme case that the target is fully occluded by other objects, our model is capable of producing complete amodal masks that closely resemble the ground truth in terms of both position and shape. Thanks to our proposed shape prior generator and coarse-to-fine module.

\begin{figure}[t]
\begin{center}
\includegraphics[width=\columnwidth]{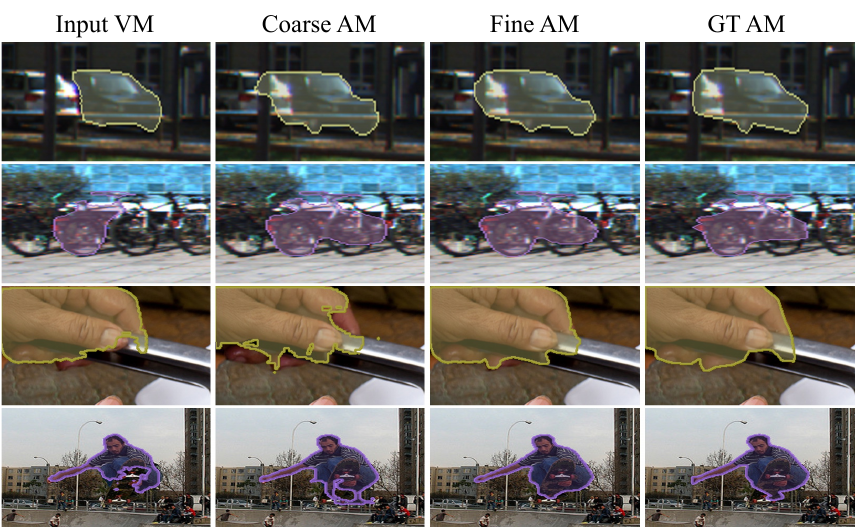}
\caption{
The coarse-to-fine progress in C2F-Seg. 
The estimated amodal mask is iteratively refined from the visible mask.
VM indicates visible mask while AM indicates amodal mask.}
\label{fig:refine_process}
\end{center}
\vskip -0.15in
\end{figure}

\begin{table}\small
\centering
\setlength{\tabcolsep}{1.5mm}{
{\begin{tabular}{lcccc}
\toprule
\multicolumn{1}{c}{\multirow{2}{*}{\textsc{Methods}}} & \multicolumn{2}{c}{KINS} & \multicolumn{2}{c}{COCOA} \\
 & mIoU$_{full}$ & mIoU$_{occ}$ & mIoU$_{full}$ & mIoU$_{occ}$  \\
\midrule
\midrule
\textit{w/o} refine  & 81.81 & 52.57     & 79.52 &  24.25   \\
single branch & 81.95 & 52.78    &  80.03 & 26.12   \\
full model   & \textbf{82.22} & \textbf{53.60}  &  \textbf{80.28} & \textbf{27.71} \\
\bottomrule
\end{tabular}}}
\caption{\textbf{Ablation results for the refine module.} We report mean-IoU metrics on KINS and COCOA to evaluate our refine module.}
\label{tab:Ablation_for_refine}
\vskip -0.1in
\end{table}

\subsection{Ablation Study}
\label{Ablation}

We further conduct ablation studies to evaluate the effectiveness of our model on both image and video datasets.

\noindent\textbf{Effect of Convolutional Refinement}.
To investigate the effectiveness of our proposed refine module, we first verified the validity of the two-branches architecture which both predicts the visible masks and amodal masks. We train C2F-Seg with the refine module only predicting the amodal masks. Further more, we conduct another experiment without the refine module. These two experiments keep the same setting as claimed before. The results are shown in Table~\ref{tab:Ablation_for_refine}. We also visualize the process that C2F-Seg predicts the precise amodal mask based on the input visible mask. In Figure~\ref{fig:refine_process}, the Coarse FMs reflect the shape prior our model has learned and the refine module improves them in detail by adding and removing some redundant regions. 
It indicates that both predict visible masks and amodal masks will help the model to distinguish and figure out the difference between the two masks. 
We can draw the conclusion that the two-branches refine module helps our model to predict amodal masks better. 

\noindent\textbf{Effect of Parameter $K$}. 
It's noteworthy that $K$ significantly influences MaskGIT~\cite{chang2022maskgit}. To investigate the effect of \(K\) on our model, we assessed various $K$ values on the COCOA and MOViD-A datasets. The results can be found in the supplementary (Table~\ref{tab:K_value}). Interestingly, the table indicates that changes in the $K$ value have only a marginal effect on our model. A potential explanation might be the difference in tasks: in our context, the module involving $K$ is designed to generate coarse masks, whereas MaskGIT focuses on RGB image production.

\begin{table}\small
\centering
\setlength{\tabcolsep}{1.5mm}{
{\begin{tabular}{ccccc}
\toprule
\multicolumn{1}{c}{\multirow{2}{*}{\textsc{GT VM}}} & \multicolumn{2}{c}{KINS} & \multicolumn{2}{c}{COCOA} \\
 & mIoU$_{full}$ & mIoU$_{occ}$ & mIoU$_{full}$ & mIoU$_{occ}$  \\
\midrule
\midrule
$\times$  & 82.22 & 53.60    &  80.28 & 27.71   \\
\checkmark  & \textbf{87.89} & \textbf{57.60}  &  \textbf{87.13} & \textbf{36.55} \\
\bottomrule
\end{tabular}}}
\caption{The Upper bound mean-IoU metrics of C2F-Seg on KINS and COCOA. It indicates that high quality visible masks help our model reach better performance.}
\label{tab:upper_bound}
\vspace{-0.15in}
\end{table}

\noindent \textbf{Upper bound of C2F-Seg}
Since our model is driven by visible mask, we try to feed it with GT visible masks to explore the upper bound metric. We keep the same setting mentioned for Image Amodal Segmentation to evaluate the best performance of C2F-Seg. We only change the predicted visible masks to GT visible masks for KINS and COCOA.
Table~\ref{tab:upper_bound} shows the mIoU metrics for the two datasets. Our model achieves 5.67/4.0 and 6.86/9.44 on KINS and COCOA respectively. It shows that our model will reach much better results with high quality visible masks.

\section{Conclusion}
In this work, we introduce a novel framework, C2F-Seg, which harnesses transformers to learn shape priors in the latent space, enabling the generation of a coarse mask. Subsequently, we deploy a dual-branch refinement module to produce an attention mask. This mask is then combined with the coarse mask and features from ResNet-50 to predict both visible and amodal masks. For video datasets, we adapt our transformer block into a spatial-temporal version to effectively capture spatio-temporal features, leading to superior amodal mask predictions.  Our model gets the precise amodal masks step by step and achieves new State-of-the-art
performance both on image and video amodal segmentation.

\noindent { \small \textbf{Acknowledgements.} This work is supported by China Postdoctoral Science Foundation (2022M710746). Yanwei
Fu is with the School of Data Science, Shanghai Key Lab of Intelligent Information Processing, Fudan University, and Fudan ISTBI-ZJNU Algorithm Centre for Brain-inspired Intelligence, Zhejiang Normal University, Jinhua, China. }

\newpage

{\small
\bibliographystyle{ieee_fullname}
\bibliography{egbib}
}

\clearpage

\appendix

\counterwithin{figure}{section}
\counterwithin{table}{section}

\section{Preliminary Knowledge}

\subsection{Detail of Vector-Quantization Module}

This module draws inspiration from the well-known VQGAN~\cite{esser2021taming}. Our aim is to reduce the learning complexity and expedite the inference process during the coarse segmentation phase. Therefore, we execute the segmentation within a low-dimensional vector-quantized latent space. Beyond what is mentioned in the main paper, the training objective is to identify the optimal compression model $\mathcal{Q}^*=\left\{E^*, G^*, \mathcal{Z}^*\right\}$, which can be expressed as:
$$
\begin{aligned}
\mathcal{Q}^*=\underset{E, G, \mathcal{Z}}{\arg \min } \max _D \mathbb{E}_{x \sim p(x)} &{\left[\mathcal{L}_{\mathrm{VQ}}(E, G, \mathcal{Z})\right.} \\
&\left.+\lambda \mathcal{L}_{\mathrm{GAN}}(\{E, G, \mathcal{Z}\}, D)\right],
\end{aligned}
$$
where
$$
\begin{aligned}
\mathcal{L}_{\mathrm{VQ}}(E, G, \mathcal{Z})=\mathcal{L}_{\mathrm{rec}} & +\left\|\operatorname{sg}[E(x)]-z_{\mathbf{q}}\right\|_2^2 \\
& +\beta\left\|\operatorname{sg}\left[z_{\mathbf{q}}\right]-E(x)\right\|_2^2
\end{aligned}
$$
and
$$
\mathcal{L}_{\mathrm{GAN}}(\{E, G, \mathcal{Z}\}, D)=[\log D(x)+\log (1-D(\hat{x}))]
$$
The adaptive weight $\lambda$ is computed as:
$$
\lambda=\frac{\nabla_{G_L}\left[\mathcal{L}_{\mathrm{rec}}\right]}{\nabla_{G_L}\left[\mathcal{L}_{\mathrm{GAN}}\right]+\delta}
$$
In this context, $\mathcal{L}_{\mathrm{rec}}$ represents the perceptual reconstruction loss~\cite{zhang2018unreasonable}. The symbol $\operatorname{sg}[\cdot]$ indicates the stop-gradient operation, while $\left\|\operatorname{sg}\left[z_{\mathbf{q}}\right]-E(x)\right\|_2^2$ is referred to as the commitment loss and has a weighting factor of $\beta$~\cite{van2017neural}. The notation $\nabla_{G_L}[\cdot]$ signifies the gradient of its input with respect to the last layer $L$ of the decoder. For numerical stability, we employ $\delta=10^{-6}$.

In our experiments, we fixed the codebook size $|\mathcal{Z}|$ at 256 across all datasets. We also omitted the attention layer from the original model. The entire iteration process for the four datasets is configured at 100k.

\subsection{Detail of Iterative Inference}

Inspired by MaskGIT~\cite{chang2022maskgit}, the mask-and-predict procedure facilitates natural sequential decoding during inference. Beginning with a token sequence that masks all amodal segments, our transformer incrementally completes the amodal segments, preserving the most confident prediction with each step. In detail, to produce a coarse mask at inference time, we commence with a blank canvas where all tokens are masked, denoted as \(Y_{\mathbf{M}}^{(0)}\) (where \(Y_{\mathbf{M}}\) represents the result after applying mask \(\mathbf{M}\) to \(Y\)). For iteration \(t\), our transformer operates as:

1. \textbf{Parallel Prediction}: Starting with the current set of masked tokens, \(Y_{\mathrm{M}}^{(t)}\), the transformer predicts the likelihoods for all masked positions at once, producing a probability matrix \(p^{(t)} \in \mathbb{R}^{N \times K}\).

2. \textbf{Token Sampling with Confidence Scoring}: At every masked location, a token is sampled based on its associated probabilities. This token's prediction score is taken as a confidence measure, showing the model's trust in its prediction. Positions that are already unmasked are automatically given full confidence, scored at \(1.0\).

3. \textbf{Dynamic Masking}: The number of tokens that should remain masked in the next iteration is computed using the mask scheduling function \(\gamma\). This accounts for the input length \(N\) and the progression of iterations \(t\) relative to the total \(T\).

4. \textbf{Update Masking Strategy}: Tokens in \(Y_{\mathbf{M}}^{(t)}\) are then updated for the next iteration. Only tokens with lower confidence scores are re-masked, as determined by a threshold value derived from the sorted confidence scores. This ensures that the transformer focuses on refining less confident tokens in the subsequent iteration.

The Iterative Inference assembles a coarse amodal mask in \(K\) steps. During each iteration, the transformer anticipates all tokens concurrently, yet retains only the most confident selections. Subsequent tokens are masked again and re-predicted in the following iteration. The mask ratio diminishes until all tokens are formulated within \(K\) iterations.

\section{Further Ablation Studies}

\subsection{Table of Ablation Study for \(K\)} 

We have carried out an ablation study to investigate the impact of \(K\) on our model. The performance of our model, across different values of \(K\), on the COCOA and MOViD-A datasets, is detailed in Table~\ref{tab:K_value}.

\begin{table}[h]\small
\centering
\setlength{\tabcolsep}{1.5mm}{
{\begin{tabular}{ccccc}
\toprule
\multicolumn{1}{c}{\multirow{2}{*}{\textsc{K}}} & \multicolumn{2}{c}{COCOA} & \multicolumn{2}{c}{MOViD-A} \\
 & mIoU$_{full}$ & mIoU$_{occ}$ & mIoU$_{full}$ & mIoU$_{occ}$  \\
\midrule
\midrule

1  & 80.16 & 27.70  &  \textbf{71.91} & \textbf{36.57} \\
2  & 80.27 & 27.68  &  71.67 & 36.30 \\
3  & 80.28 & \textbf{27.71}  &  71.67 & 36.13 \\
5  & 80.28 & 27.60  &  71.58 & 35.88 \\
8  & \textbf{80.31} & 27.57  &  71.46 & 35.53 \\
10  & 80.24 & 27.28  &  71.42 & 35.60 \\
12  & 80.27 & 27.44  &  71.41 & 35.44 \\

\bottomrule
\end{tabular}}}
\caption{Ablation results for $K$ on COCOA and MOViD-A.}
\label{tab:K_value}
\vspace{-0.1in}
\end{table}

In order to further evaluate the effectiveness of our model both on image and video datasets, we conduct the following two experiments.

\subsection{Effect of Time Rolling in Transformer}

We also investigate the effectiveness of Spatial Temporal(ST) module used in our video version of C2F-Seg.
The ST module is proposed in~\cite{gupta2022maskvit} and we modify the module with an extra roll mechanism which will help C2F-Seg to model the whole video, and make full use of transformer to extract spatiotemporal information features over long distances.
In this part, we evaluate the effect of each module.
We train our model with full ST module, without ST module, and without roll mechanism respectively on the two video datasets. The results are shown in Table \ref{tab:abalation_fishbowl_MOVi_amodal}. 
Results indicate the effectiveness of the ST module as well as our introduced roll mechanism.

\subsection{The Effect of Attention Mechanism in Refinement Module}

To investigate the effectiveness of the attention calculated in our proposed refine module, we train C2F-Seg with and without calculating attention separately on KINS and COCOA. Table~\ref{tab:Ablation_for_attn} shows the mIoU metrics for the two datasets. The results indicate our attention mechanism improves the quality of amodal masks.

\begin{table} \small
\centering
\setlength{\tabcolsep}{1.2mm}{
{\begin{tabular}{lcccc}
\toprule
\multicolumn{1}{c}{\multirow{2}{*}{\textsc{Methods}}} & \multicolumn{2}{c}{Fishbowl} & \multicolumn{2}{c}{MOViD-A} \\
 & mIoU$_{full}$ & mIoU$_{occ}$ & mIoU$_{full}$ & mIoU$_{occ}$  \\
\midrule
\midrule
\textit{w/o} ST module & 89.64   & 78.93    & 67.19   & 26.48 \\
\textit{w/o} roll      & 90.91   & 80.01   & 69.92   & 32.35  \\
full model   & \textbf{ 91.68} & \textbf{81.21} &  \textbf{71.67} & \textbf{36.13}  \\
\bottomrule
\end{tabular}}}
\caption{\textbf{Ablation results for our STTB module for Video task.} We report the mean-IoU metric for Fishbowl and MOViD-A to evaluate our design for spatio-temporal feature.} 
\label{tab:abalation_fishbowl_MOVi_amodal}
\end{table}

\begin{table}\small
\centering
\setlength{\tabcolsep}{1.5mm}{
{\begin{tabular}{lcccc}
\toprule
\multicolumn{1}{c}{\multirow{2}{*}{\textsc{Methods}}} & \multicolumn{2}{c}{KINS} & \multicolumn{2}{c}{COCOA} \\
 & mIoU$_{full}$ & mIoU$_{occ}$ & mIoU$_{full}$ & mIoU$_{occ}$  \\
\midrule
\midrule
\textit{w/o attn}   & 82.07 & 52.98     & 80.15 &  26.85   \\
\textit{w. attn}    & \textbf{82.22} & \textbf{53.60}  &  \textbf{80.28} & \textbf{27.71} \\
\bottomrule
\end{tabular}}}
\caption{\textbf{Ablation results for the attention mechanism.} Mean-IoU metrics on KINS and COCOA to evaluate this mechanism.}
\label{tab:Ablation_for_attn}
\vskip -0.1in
\end{table}

\section{Supports for the claim of shape prior}
\begin{figure*}[h]
\vskip -0.14in
\centering
\includegraphics[width=15cm]{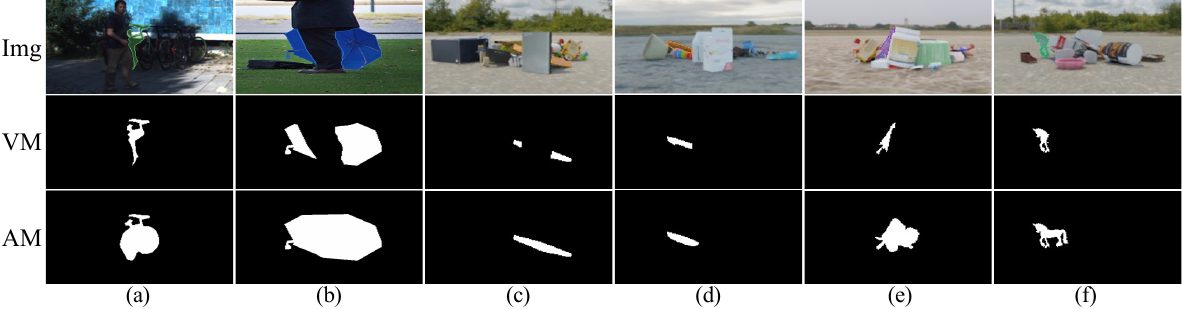}
\vskip -0.05in
\caption{
\small Supports for shape prior. \label{fig:shape_prior}
}
\vskip -0.05in
\end{figure*}

Our claim of shape prior is based on a common phenomenon, which is supported by Figure~\ref{fig:shape_prior} showcasing six randomly selected cases.
In the figure, the arrangement from top to bottom includes the images, the visible masks, and the amodal masks. Specifically, (a) is from KINS, (b) is from COCOA, and the remaining cases are from MOViD-A.
We can observe that:

\textbf{(1)} The visible masks of these cases exhibit significant differences compared to their corresponding amodal masks due to occlusion caused by different poses. 

\textbf{(2)} Besides, viewpoint variations may lead to differences in the shape prior. This is exemplified by cases (b)-(d), where the shape prior differs from the original in regular view.

\section{More Qualitative Results}

In order to more intuitively illustrate the strengths of our algorithm, and to compare it with the baselines, we select KINS and MOViD-A to show more qualitative results to demonstrate the effectiveness of our method.

\subsection{Visualization on KINS Dataset}
\begin{figure*}[h]
\vskip -0.1in
\begin{center}
\includegraphics[width=16cm]{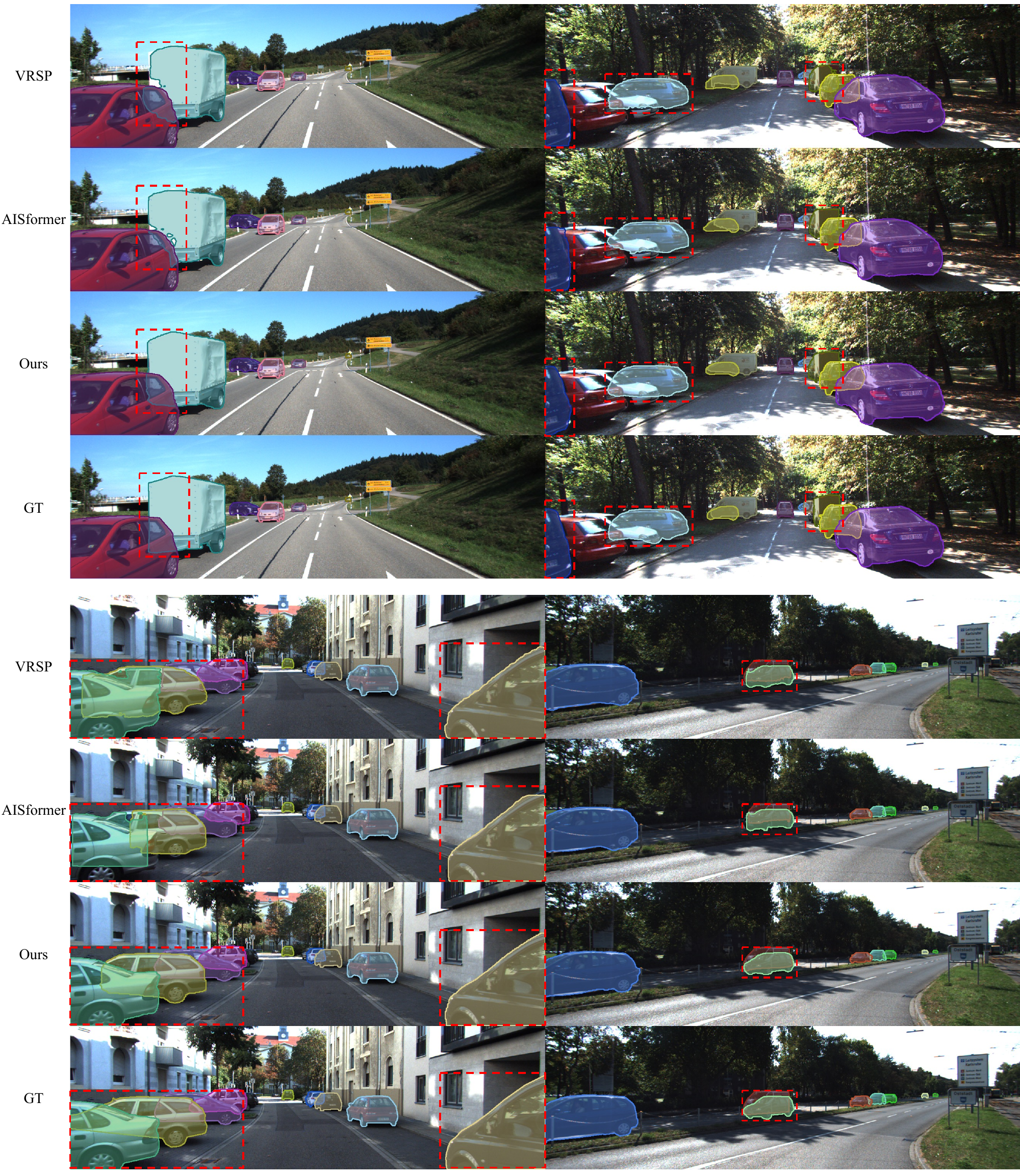}
\caption{
The qualitative results estimated by VRSP, AISFormer, and our method. GT indicates ground-truth amodal mask.
}
\label{fig:KINS_all}
\end{center}
\vskip -0.3in
\end{figure*}

To show the performance of our method on real scenarios, we show more results from KINS in Figure~\ref{fig:KINS_all}. In these images, for fair comparison, we select the intersection of the amodal masks predicted by VRSP~\cite{xiao2021amodal} and AISFormer~\cite{tran2022aisformer}. Our algorithm completes the occluded cars better than all the baselines on KINS, which will help to improve the safety of autonomous driving significantly if applied to real scenarios.

\subsection{Visualization on MOViD-A Dataset}
\begin{figure*}[h]
\begin{center}
\includegraphics[width=9cm]{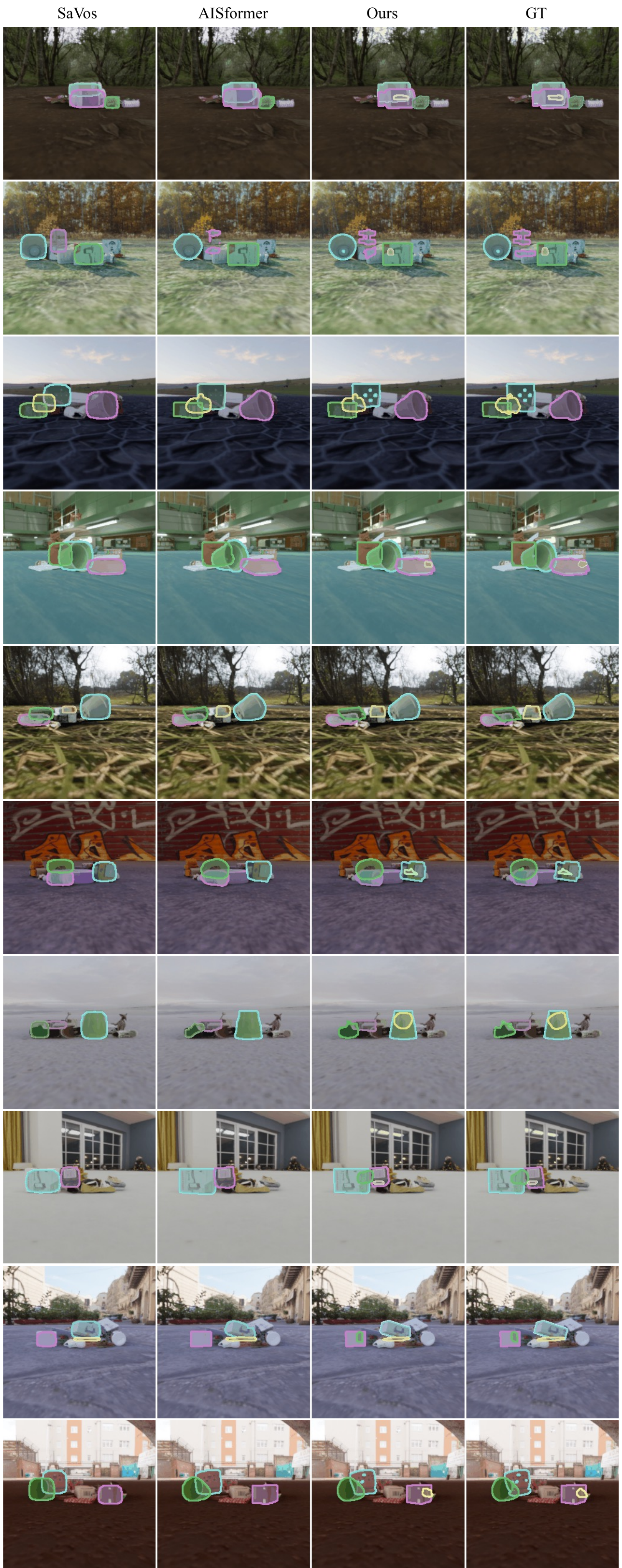}
\caption{
The qualitative results estimated by SaVos, AISFormer, and our method. GT indicates ground-truth amodal mask.
}
\label{fig:MOViD-A_vis}
\end{center}
\vskip -0.1in
\end{figure*}

We show the qualitative results estimated by the best baseline video and image-based amodal method on MOViD-A respectively in Figure~\ref{fig:MOViD-A_vis}. Our method predicts the invisible masks excellently by extracting valid spatio-temporal features and outperforms all the baselines.

\section{Limitations and Future Works}
We propose a coarse-to-fine framework that leverages shape prior for amodal segmentation. Despite it has achieved significant advantages in both image and video-based benchmarks, our proposed C2F-Seg still faces several limitations. One is the additional input of the pre-detected visible mask. It is essential but not efficient, since we need to specify the target when multiple objects occur in the same scene. In future work, we will either replace it with a single point or incorporate our framework with an end-to-end detection branch, to effectively decrease the input requirement. Another limitation may lie in objects which are heavily or fully occluded. Though our introduced Spatial Temporal Transformer Block successfully mitigates this problem by aggregating multi-frame shape priors, amodal masks of some frames are not precise due to the ill-posed problem. We will explicitly design modules to utilize spatio-temporal prior and constraint the consistence of masks between adjacent frames.

\end{document}